\documentclass[letterpaper,10pt,conference]{ieeeconf}  
\IEEEoverridecommandlockouts

\usepackage{amsmath,amssymb,amsfonts}
\usepackage{bm}
\usepackage{rigidnotation}
\SetConciseNotation{\BooleanTrue}
\usepackage[dvipsnames]{color, colortbl, xcolor}
\usepackage{siunitx}
\usepackage{graphicx}
\usepackage{hyperref}
\usepackage{balance}
\usepackage{pdfpages}
\usepackage{pgfplots}
\pgfplotsset{compat=1.17}
\tikzset{font={\fontsize{8pt}{8}\selectfont}} 
\usepackage{array} 
\usepackage{longtable} 
\usepackage{multirow} 
\usepackage{tabularx}
\usepackage{float} 
\usepackage{subcaption}
\usepackage{booktabs} 
\usepackage{glossaries}
\setacronymstyle{long-short}

\newacronym{nn}{NN}{nearest neighbor}
\newacronym{mnn}{MNN}{mutual nearest neighbor}

\newacronym{pose}{pose}{position and orientation}

\newacronym{iqr}{IQR}{interquartile range}
\newacronym{mad}{MAD}{median absolute deviation}

\newacronym{mlp}{MLP}{multilayer perceptron}
\newacronym{tp}{TP}{true positives}
\newacronym{fp}{FP}{false positives}
\newacronym{fn}{FN}{false negatives}
\newacronym{tn}{TN}{true negatives}
\newacronym{mma}{MMA}{mean matching accuracy}
\newacronym{mscore}{M. Score}{matching score}
\newacronym{map}{mAP}{mean average precision}
\newacronym{ace}{ACE}{average corner error}

\newacronym{nms}{NMS}{non-maximum suppression}
\newacronym{ransac}{RANSAC}{random sample consensus}
\newacronym{lma}{LMA}{the Levenberg–Marquardt algorithm}
\newacronym{dlt}{DLT}{direct linear transformation}
 
\newacronym{relu}{ReLU}{rectified linear unit}
\newacronym{zncc}{ZNCC}{zero-normalized cross-correlation}
\newacronym{iou}{IoU}{intersection-over-union}
\newacronym{mse}{MSE}{mean squared error}

\newacronym{svd}{SVD}{singular value decomposition}
\newacronym{nls}{NLS}{nonlinear least squares}

\newacronym{uav}{UAV}{unmanned aerial vehicle}
\newacronym{ugv}{UGV}{unmanned ground vehicle}
\newacronym{dof}{DOF}{degree-of-freedom}

\newacronym{vio}{VIO}{visual-inertial odometry}
\newacronym{tio}{TIO}{thermal-inertial odometry}
\newacronym{to}{TO}{thermal odometry}
\newacronym{imu}{IMU}{inertial measurement unit}
\newacronym{lidar}{LiDAR}{light detection and ranging}
\newacronym{slam}{SLAM}{simultaneous localization and mapping}
\newacronym{sfm}{SFM}{stucture from motion}

\newacronym{ir}{IR}{infrared}
\newacronym{lwir}{LWIR}{long-wave infrared}
\newacronym{mwir}{MWIR}{mid-wave infrared}
\newacronym{nir}{NIR}{near infrared}
\newacronym{fir}{FIR}{far infrared}
\newacronym{swir}{SWIR}{short-wave infrared}
\newacronym{tir}{TIR}{thermal infrared}
\newacronym{vis}{VIS}{visible-spectrum}
\newacronym{emr}{EMR}{electromagnetic radiation}

\newacronym{pc}{PC}{phase congruency}
\newacronym{mi}{MI}{mutual information}
\newacronym{nmi}{NMI}{normalized mutual information}
\newacronym{ncc}{NCC}{normalized cross-correlation}
\newacronym{ssim}{SSIM}{Structural Similarity Index}
\newacronym{dssim}{DSSIM}{Structural Dissimilarity Index}
\newacronym{sad}{SAD}{sum of absolute differences}
\newacronym{ssd}{SSD}{sum of squared differences}

\newacronym[]{rmse}{RMSE}{root mean square error}

\newacronym{hopc}{HOPC}{histogram of oriented phase congruency}
\newacronym{cfog}{CFOG}{channel features of oriented gradients}
\newacronym{hog}{HOG}{histogram of gradients}
\newacronym{rift}{RIFT}{Radiation-Invariant Feature Transform}
\newacronym{rift2}{RIFT2}{Radiation-Invariant Feature Transform 2}

\newacronym{nuc}{NUC}{non-uniformity correction}
\newacronym{ffc}{FFC}{flat-field correction}

\newacronym{ros}{ROS}{Robot Operating System}

\newacronym{eoh}{EOH}{Edge-Oriented Histogram}
\newacronym{mfd}{MFD}{Multispectral Feature Descriptor}
\newacronym{lghd}{LGHD}{Log-Gabor Histogram Descriptor}
\newacronym{cmm-net}{CMM-Net}{Cross-Modality Matching Network}
\newacronym{cvc}{CVC}{Computer Vision Center}
\newacronym{cats}{CATS}{Colour and Thermal Stereo}

\newacronym{lift}{LIFT}{Learned Invariant Feature Transform}
\newacronym{r2d2}{R2D2}{Reliable and Repeatable Detector and Descriptor}
\newacronym{lf-net}{LF-Net}{Local Feature Network}
\newacronym{d2-net}{D2-Net}{Detect-and-Describe Network}
\newacronym{redfeat}{ReDFeat}{Recoupling Detection and Description for Multimodal Feature Learning}

\newacronym{surf}{SURF}{Speeded Up Robust Features}
\newacronym{fast}{FAST}{Features from Accelerated Segment Test}
\newacronym{sift}{SIFT}{Scale-Invariant Feature Transform}
\newacronym{orb}{ORB}{Oriented FAST and Rotated BRIEF}
\newacronym{brisk}{BRISK}{Binary Robust Invariant Scalable Keypoints}
\newacronym{brief}{BRIEF}{Binary Robust Independent Elementary Features}
\newacronym{freak}{FREAK}{Fast Retina Keypoint}
\newacronym{dog}{DoG}{difference of Gaussians}
\newacronym{alike}{ALIKE}{Accurate and Lightweight Keypoint Detection and Descriptor Extraction}

\newacronym{ekf}{EKF}{Extended Kalman Filter}

\newacronym{cnn}{CNN}{convolutional neural network}
\newacronym[plural=GANs, firstplural=generative adversarial networks (GAN)]{gan}{GAN}{generative adversarial network}
\newacronym[plural=cGANs, firstplural= conditional generative adversarial networks (cGAN)]{cgan}{cGAN}{conditional generative adversarial network}
\newacronym[plural=CycleGANs, firstplural= cycle-consistent adversarial networks (CycleGANs)]{cyclegan}{CycleGAN}{cycle-consistent adversarial network}
\newacronym{vit}{ViT}{vision transformer}

\newacronym{jpl}{JPL}{Jet Propulsion Laboratory}

\newacronym{vtr}{VT\&R}{Visual Teach and Repeat}

\usepackage[style=ieee,citestyle=numeric-comp,giveninits=true,maxnames=6, minnames=1, maxcitenames=2, mincitenames=1, backend=bibtex,doi=false,isbn=false,url=false,eprint=false]{biblatex}
\AtBeginBibliography{\footnotesize}
\AtEveryBibitem{
	\clearfield{issn} 
	\clearfield{ibsn} 
	\clearfield{note} 
	\clearfield{addendum} 
	\clearlist{publisher} 
	\clearlist{language} 
	\clearfield{editor} 
}
\DefineBibliographyStrings{english}{andothers = {\textup{et al.\ }},}

\DeclareMathAlphabet{\mbf}{OT1}{ptm}{b}{n}

\newcommand{\norm}[1]{\left\Vert#1\right\Vert}
\newcommand{\abs}[1]{\left\vert#1\right\vert}
\newcommand{\set}[1]{\left\{#1\right\}}
\newcommand{\Real}{\mathbb R}
\newcommand{\bbm}{\begin{bmatrix}}
\newcommand{\ebm}{\end{bmatrix}}
\newcommand{\mbs}[1]{{\boldsymbol{#1}}}

\newcommand{\hatmbf}[1]{\hat{\mbf{#1}}}

\NewRigidNotation{NeuronFun}{f_\textnormal{neuron}}
\NewRigidNotation{NumNeuronInput}{N}
\NewRigidNotation{NeuronInput}{\mbf{x}}
\NewRigidNotation{NeuronInputElement}{{x}}
\NewRigidNotation{ActivationFun}{h}
\NewRigidNotation{NeuronWeights}{\mbf{w}}
\NewRigidNotation{NeuronWeightElement}{{w}}
\NewRigidNotation{NeuronBias}{b}

\NewRigidNotation{LayerFun}{\mbf{f}}
\NewRigidNotation{LayerFunNamed}{\mbf{f}_\textnormal{layer}}
\NewRigidNotation{LayerInput}{\mbf{x}_l}
\NewRigidNotation{LayerActivationFun}{\mbf{h}}
\NewRigidNotation{LayerWeights}{\mbf{W}_l}
\NewRigidNotation{LayerBias}{\mbf{b}_l}
\NewRigidNotation{LayerParams}{\mbs{\theta}_{l}}

\NewRigidNotation{LayerIndex}{l}
\NewRigidNotation{NumLayerOutput}{N_l}
\NewRigidNotation{NumLayers}{L}
\NewRigidNotation{NumNetworkOutput}{N_L}

\NewRigidNotation{NetworkOutputVector}{\hatmbf{y}}
\NewRigidNotation{NetworkOutputElement}{\hat{y}}
\NewRigidNotation{NetworkFun}{\mbf{f}_\textnormal{net}}
\NewRigidNotation{NetworkInput}{\mbf{x}}
\NewRigidNotation{NetworkParams}{\mbs{\theta}}

\NewRigidNotation{OneHotVector}{\mbf{y}}
\NewRigidNotation{OneHotElement}{y}
\NewRigidNotation{LossSquaredError}{\Loss{\mathrm{SE}}}

\NewRigidNotation{SoftmaxVector}{\mbs{\sigma}}
\NewRigidNotation{SoftmaxElement}{{\sigma}}

\NewRigidNotation{LossCrossEntropy}{\Loss{\mathrm{CE}}}

\NewRigidNotation{DetLossWeight}{\mbs{\rho}}
\NewRigidNotation{DetLossWeightElement}{{\rho}}

\NewRigidNotation{LossAvg}{\bar{\Loss}}
\NewRigidNotation{LayerParamsMin}{\mbs{\theta}^*}
\NewRigidNotation{SingleParam}{\theta}
\NewRigidNotation{LearningRate}{\alpha}

\NewRigidNotation{NumBatch}{N_\textnormal{B}}

\NewRigidNotation{ChannelOut}{\beta}
\NewRigidNotation{ChannelIn}{\alpha}
\NewRigidNotation{SetDisplacement}{\mathcal{D}}

\NewRigidNotation{CNNDimIn}{N_{l-1}}
\NewRigidNotation{CNNDimOut}{N_{l}}
\NewRigidNotation{CNNDimKernel}{K}
\NewRigidNotation{Padding}{P}
\NewRigidNotation{Stride}{S}

\NewRigidNotation{FastNum}{N}
\NewRigidNotation{FastThresh}{t}

\NewRigidNotation{NumMatch}{N_\textnormal{M}}

\NewRigidNotation{HomVector}{\mbf{h}}
\NewRigidNotation{HomVectorDelta}{\bm{\Delta}\HomVector}
\NewRigidNotation{HomVectorDeltaBest}{\bm{\Delta}\HomVector^*}
\NewRigidNotation{HomVectorInit}{\HomVector[\hat]}
\NewRigidNotation{MatchMatrix}{\mbf{A}}
\NewRigidNotation{MatchWeightMatrix}{\mbf{W}}
\NewRigidNotation{MatchWeightElement}{w}

\NewRigidNotation{FunHom}{\mbf{f}}
\NewRigidNotation{ErrorNLS}{E_{NLS}}
\NewRigidNotation{ErrorRes}{\mbf{r}}
\NewRigidNotation{ErrorJac}{\mbf{J}}
\NewRigidNotation{ErrorVarA}{\mbf{A}}
\NewRigidNotation{ErrorVarB}{\mbf{b}}
\NewRigidNotation{ErrorVarC}{c}
\NewRigidNotation{NLSStepSize}{\alpha}

\NewRigidNotation{LMADamp}{\lambda}
\NewRigidNotation{FunDiag}{\textnormal{diag}}
\NewRigidNotation{Eye}{\mbf{I}}

\NewRigidNotation{RepeatThresh}{\epsilon_R}
\NewRigidNotation{RepeatFun}{\mathrm{R}}
\NewRigidNotation{MatchSet}{\mathcal{M}}
\NewRigidNotation{MatchThresh}{\epsilon_M}

\NewRigidNotation{TruePositive}{\textnormal{TP}}
\NewRigidNotation{FalsePositive}{\textnormal{FP}}
\NewRigidNotation{TrueNegative}{\textnormal{TN}}
\NewRigidNotation{FalseNegative}{\textnormal{FN}}

\NewRigidNotation{Hom}{\mbf{H}}
\newcommand{\HomGt}{\Hom{s}{}{t}}
\newcommand{\HomEst}{\Hom[\hat]{s}{}{t}}
\NewDocumentCommand\hatbar{m}{\hat{\bar{#1}}}

\newcommand{\HomGtNormSimp}{\Hom[\bar]{}{}{}}
\newcommand{\HomEstNormSimp}{\Hom[\hatbar]{}{}{}}
\newcommand{\HomRes}{\HomGtNormSimp^{-1}\HomEstNormSimp}
\newcommand{\HomResInv}{\HomEstNormSimp^{-1}\HomGtNormSimp}

\NewRigidNotation{FeatureMap}{\mbf{F}}
\NewRigidNotation{DescMap}{\mbf{D}}
\NewRigidNotation{LogitMap}{\mbf{K}}
\NewRigidNotation{LogitCell}{\mbf{k}}
\NewRigidNotation{LogitCellElement}{{k}}
\NewRigidNotation{LogitMapMod}{\LogitMap[\tilde]}
\NewRigidNotation{LogitCellElementMod}{\LogitCellElement[\tilde]}
\NewRigidNotation{Window}{\mathcal{W}}
\NewRigidNotation{ProbMap}{\mbf{P}}
\NewRigidNotation{ProbCell}{\mbf{p}}
\NewRigidNotation{ProbCellElement}{{p}}

\NewRigidNotation{Point}{\mbf{q}}
\newcommand{\PointSrc}{\Point{}{}{s}}
\newcommand{\PointTrg}[1][]{\Point[#1]{}{}{t}}
\newcommand{\PointTrgPseudo}{\Point[\hat]{}{}{t}}

\NewRigidNotation{PointRealWorld}{\mbf{p}}  
\NewRigidNotation{IntrinsicMatrix}{\mbf{K}}  
\NewRigidNotation{Focal}{f}  
\NewRigidNotation{Central}{c}

\newcommand{\NumTrgKp}{N_\textnormal{KT}}
\newcommand{\NumKp}{N_\textnormal{K}}

\NewRigidNotation{PointSet}{\mbf{Q}}

\newcommand{\PointSetSrcNorm}{\PointSet[\bar]{}{}{s}}
\newcommand{\PointSetTrgPseudoNorm}{\PointSet[\hatbar]{}{}{t}}

\NewRigidNotation{Desc}{\mbf{d}}
\newcommand{\DescSrc}{\Desc{}{}{s}}
\newcommand{\DescTrg}{\Desc{}{}{t}}
\newcommand{\DescTrgPseudo}{\Desc[\hat]{}{}{t}}

\NewRigidNotation{VecAElement}{x}
\NewRigidNotation{VecA}{\mbf{x}}
\NewRigidNotation{VecAAvg}{\bar{x}}
\NewRigidNotation{VecBElement}{y}
\NewRigidNotation{VecB}{\mbf{y}}
\NewRigidNotation{VecBAvg}{\bar{y}}

\NewRigidNotation{Score}{s}
\newcommand{\ScoreKP}{\Score{\mathrm{K}}{}{}}
\newcommand{\ScoreKPSource}{\Score{\mathrm{K}}{}{s}}
\newcommand{\ScoreKPTargetPseudo}{\Score[\hat]{\mathrm{K}}{}{t}}
\newcommand{\ScoreMatch}{\Score{\mathrm{M}}{}{}}
\newcommand{\ScoreInlier}{\Score{\mathrm{I}}{}{}}

\newcommand{\Temperature}{\tau}
\newcommand{\fzncc}{f_{\mathrm{zncc}}}
\NewRigidNotation{ZnccWeightElement}{\phi}
\NewRigidNotation{ZnccWeightVector}{\mbs{\phi}}
\NewRigidNotation{ZnccWeightMatrix}{\mbs{\Phi}}
\newcommand{\MatchError}{x}
\newcommand{\OutlierThreshold}{a}
\newcommand{\OutlierSharpness}{b}

\NewRigidNotation{Identity}{\mbf{I}}
\NewRigidNotation{Error}{\mbf{R}}
\NewRigidNotation{ErrorElement}{r}
\newcommand{\ErrorCorner}{\Error{\mathrm{C}}}
\newcommand{\PointsCornerNorm}{\PointSet[\bar]{\mathrm{C}}}
\newcommand{\ErrorFrob}{\Error{\mathrm{F}}}
\newcommand{\ErrorTrans}{\Error{\mathrm{T}}}
\newcommand{\ErrorInv}{^{\hspace{0.1em}'}}

\newcommand{\LossCorner}{\Loss{\mathrm{C}}}
\newcommand{\LambdaCorner}{\lambda_{\mathrm{C}}}
\newcommand{\LossFrob}{\Loss{\mathrm{F}}}
\newcommand{\LambdaFrob}{\lambda_{\mathrm{F}}}
\newcommand{\LossTrans}{\Loss{\mathrm{T}}}
\newcommand{\LambdaTrans}{\lambda_{\mathrm{T}}}

\newcommand{\frobust}{f_{\mathrm{robust}}}
\newcommand{\RobustError}{\ErrorElement}
\NewRigidNotation{RobustThreshold}{c}

\NewRigidNotation{Loss}{\mathcal{L}}

\NewRigidNotation{Correspondence}{g}

\newcommand{\DescCellSrc}{\Desc{}{}{s}}
\newcommand{\DescCellTrg}{\Desc{}{}{t}}

\newcommand{\MarginPos}{m_\mathrm{P}}
\newcommand{\MarginNeg}{m_\mathrm{N}}
\newcommand{\LossDesc}{\Loss{\mathrm{D}}}
\newcommand{\LambdaDesc}{\lambda_{\mathrm{D}}}
\newcommand{\LambdaPos}{\lambda_{\mathrm{P}}}
\newcommand{\LossDetector}{\Loss{\mathrm{K}}}
\newcommand{\LambdaDetector}{\lambda_\mathrm{K}}

\NewRigidNotation{Quartile}{Q}
\NewRigidNotation{Quantile}{Q}

\NewRigidNotation{determinantthresh}{\eta}

\newcommand{\PointsCornerFull}{\PointSet{\mathrm{C}}}
\NewRigidNotation{ACE}{{e}}
\NewRigidNotation{ACEThresh}{\epsilon}
\NewRigidNotation{ACESet}{\mathcal{E}}

\NewRigidNotation{RansacWeight}{\omega}

\title{\LARGE \bf Learning Cross-Spectral Point Features with Task-Oriented Training}

\author{Mia Thomas, Trevor Ablett, and Jonathan Kelly$^\dagger$
	\thanks{All authors are with the STARS Laboratory at the University of Toronto Institute for Aerospace Studies, Toronto, Ontario, Canada. {\tt\footnotesize <firstname>.<lastname>@robotics.utias.utoronto.ca}}
	\thanks{$^\dagger$Jonathan Kelly is a Vector Institute for Artificial Intelligence Faculty Affiliate.}%
}

\begin{document}

\maketitle
\thispagestyle{empty}
\pagestyle{empty}

\begin{abstract}
Unmanned aerial vehicles (UAVs) enable operations in remote and hazardous environments, yet the visible-spectrum, camera-based navigation systems often relied upon by UAVs struggle in low-visibility conditions.
Thermal cameras, which capture long-wave infrared radiation, are able to function effectively in darkness and smoke, where visible-light cameras fail.
This work explores learned cross-spectral (thermal-visible) point features as a means to integrate thermal imagery into established camera-based navigation systems.
Existing methods typically train a feature network's detection and description outputs directly, which often focuses training on image regions where thermal and visible-spectrum images exhibit similar appearance.
Aiming to more fully utilize the available data, we propose a method to train the feature network on the tasks of matching and registration.
We run our feature network on thermal-visible image pairs, then feed the network response into a differentiable registration pipeline.
Losses are applied to the matching and registration estimates of this pipeline.
Our selected model, trained on the task of matching, achieves a registration error (corner error) below 10 pixels for more than 75\% of estimates on the MultiPoint dataset. 
We further demonstrate that our model can also be used with a classical pipeline for matching and registration.

\end{abstract}


\glsresetall 

\section{Introduction}
\label{sec:introduction}

\begin{figure}[t]
 \centering
 \includegraphics[width=3.3in, trim=4in 0.25in 2.5in 2.79in, clip]{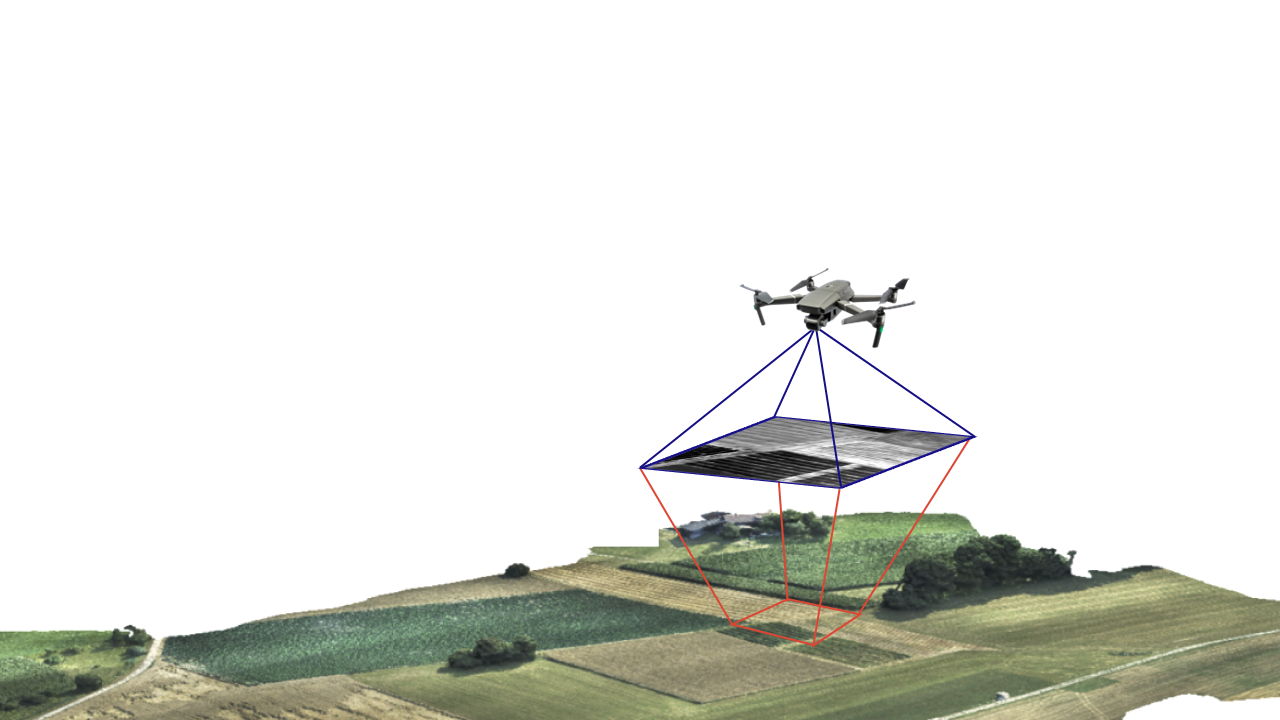}
 \caption{Conceptual rendering of a \acrshort*{uav} using an onboard thermal camera to navigate within a visible-spectrum map. Images from MultiPoint dataset~\cite{achermann_multipoint_2020}.}
 \label{fig:motivation}
 \vspace{-10pt}
\end{figure}

\Glspl*{uav} play a vital role in a wide range of applications such as search and rescue, wildlife monitoring, firefighting, agriculture, and building inspection~\cite{couturier_review_2021}.
\Glspl*{uav} often rely on visual (i.e., camera-based) navigation systems to operate in GPS-denied environments.
Currently, these systems primarily use visible-spectrum (also referred to as \textit{visible} or RGB) cameras, which face performance limitations in low-visibility environments.
This is particularly restrictive for \glspl*{uav}, as weight and power constraints often rule out onboard lighting.
Thermal cameras, however, operate effectively in dark and obscured environments. 
Using thermal imaging in concert with existing visible-spectrum systems could extend \gls*{uav} operations to previously untenable conditions.
As illustrated in Figure~\ref{fig:motivation}, this combined approach could enable thermal camera-equipped \glspl*{uav} to navigate using RGB maps during poor visibility, potentially leveraging existing RGB datasets and georeferenced images (e.g., from Google Earth~\cite{noauthor_google_nodate}).

Cross-spectral (thermal-visible in our case) point features offer a light-weight and versatile tool to integrate thermal imagery into RGB-based visual navigation systems.
A point feature---composed of a 2D keypoint (also known as a detection) and an associated descriptor vector---is a building block for a range of visual navigation systems such as pose estimation, place recognition, and scene reconstruction.

Learned point features have demonstrated superior accuracy and speed compared to handcrafted algorithms at tackling large viewpoint and appearance changes in the visible domain \cite{detone_superpoint_2018, dusmanu_d2-net_2019, revaud_r2d2_2019} and have shown promise in overcoming the appearance difference or \textit{domain gap} between thermal and RGB images \cite{achermann_multipoint_2020,deng_redfeat_2022}.
However, since point features lack a strict definition, crafting a supervisory signal is challenging, particularly for learning cross-spectral features. 
Existing supervision approaches concentrate training on image regions with smaller domain gaps.
However, a large amount of appearance variation across spectra is to be expected and, ideally, accounted for in training.

We alternatively propose a task-oriented approach to supervision.
That is, we train our feature network on the tasks of matching and registration (homography estimation in our case), aiming to `allow' our network to identify useful features for these tasks, even across large domain gaps.
We train multiple model variations with different task-based losses and evaluate the performance of our method, alongside a set of baselines, on multiple matching and registration pipelines.
This work makes the following contributions: 
\begin{itemize}
\item a comparison of different task-based loss formulations; 
\item a feature model that achieves lower registration error on the MultiPoint dataset compared to a set of baseline methods; and
\item experiments that demonstrate our model's compatibility with a classical pipeline.
\end{itemize}
Overall, we show that our task-based approach to learning cross-spectral point features can improve registration performance without limiting our model to use with a single pipeline.

\section{Related Work}
\label{sec:related-work}

This section presents an overview of work on cross-spectral point feature methods, covering handcrafted approaches, local learned features, and dense learned features.
 
Traditional, visible-spectrum algorithms typically extract point features 
either from local image gradients, such as the \gls*{sift} algorithm \cite{lowe_distinctive_2004}, or binary pixel-to-pixel comparisons in a local image patch, as in the \gls*{orb} algorithm \cite{rublee_orb_2011}.
These approaches, along with other visible-spectrum methods, often fail when faced with the nonlinear intensity variations characteristic of cross-spectral data \cite{cronje_comparison_2012}.

Handcrafted methods developed for cross-spectral data often rely on the fact that contours are generally preserved between visible-spectrum and thermal images \cite{morris_statistics_2007}.
For example, \textcite{aguilera_multispectral_2012} first extract contours using a Canny edge detector \cite{canny_computational_1986}, then apply Sobel filters to compute a histogram of gradient orientations around each keypoint. 
Similarly, a handful of works construct feature descriptors from the responses of high-frequency log-Gabor filters \cite{nunes_local_2017, aguilera_lghd_2015, li_rift_2020}. 
While an improvement on the naive use of visible-spectrum methods, these feature extraction processes are somewhat slow, taking more than one second per image \cite{nunes_local_2017}, and their accuracy could be improved.

Learning has emerged as the dominant paradigm to bridge the cross-spectral domain gap: learned methods, in addition to being faster than handcrafted approaches, have proven more accurate under dramatic appearance and viewpoint changes \cite{detone_superpoint_2018, dusmanu_d2-net_2019, revaud_r2d2_2019,achermann_multipoint_2020,deng_redfeat_2022}.
Patch-based or \textit{local} learned methods replace the descriptor algorithm, which encodes a unique vector from the patch around a keypoint, with a \gls*{cnn}.
For example, \cite{aguilera_cross-spectral_2017} trains a quadruplet network, or Q-Net, on sets of aligned thermal and visible-spectrum image patches using a contrastive loss, which simultaneously encourages similarity between descriptors of the same patch and distinctiveness from descriptors of non-corresponding patches.
Other approaches, such as \cite{jeong_learning_2019}, jointly train an image translation network and a descriptor network.
These local learned methods, however, remain dependent on handcrafted detectors and operate on low-level image information, which is unreliable for large domain gaps.

Recent work focuses on \textit{dense} learned features that operate on whole images, which use high-level information as a means to overcome the lack of a relationship between individual pixel values across spectra.
These methods typically encode the input image into a latent representation that branches into separate decoders for detection and description. 
Further, dense learned methods often employ a contrastive loss for descriptor training.

In MultiPoint \cite{achermann_multipoint_2020}, the authors generate pseudo-ground truth for the keypoints of aligned thermal and visible-spectrum image pairs using a process known as multi-spectral homographic adaptation. 
This process aggregates the responses of a base detector on warped versions of both images into a single heatmap.
This method of supervision is limited by the robustness of the base detector.
We use the framework of \cite{achermann_multipoint_2020} as a starting point for our method.

A number of works build on MultiPoint: 
\cite{elsaeidy_infrared--optical_2022} incorporates a thermal-to-visible image translation network and
\cite{elsaeidy_swin_2023} replaces the \gls*{cnn} encoder with a transformer-style architecture.
XPoint~\cite{yagmur_xpoint_2024} introduces a number of changes to~\cite{achermann_multipoint_2020}:
the authors use a visual state space model for the encoder 
and a different base detector for homographic adaptation, swapping the original visible-spectrum detector for a handcrafted cross-spectral detector.
XPoint further adds a weighting to the detector loss. 
Finally, the authors add a third decoder that regresses a homography estimate from the encoder output. 
However, unlike our method, this homography estimation head lacks a surrounding geometric framework.

A competing cross-spectral method, ReDFeat \cite{deng_redfeat_2022}, drives detection by encouraging peaks within local windows and employing an edge-based prior.
The authors then apply a similarity loss between the network responses to each spectra. 
However, in the absence of an external signal, the network risks collapsing to a degenerate output (e.g., a uniform response), requiring the authors to couple the detection and description losses in a way that masks areas of low matching confidence. 
While \cite{deng_redfeat_2022} attempts to reduce the use of handcrafted algorithms for supervision compared to \cite{achermann_multipoint_2020},
we instead choose to augment the approach of \cite{achermann_multipoint_2020} with our task-oriented training.

\section{Methodology}
\label{sec:methodology}

\begin{figure*}[t]
	\centering
	\includegraphics[width=6.9in]{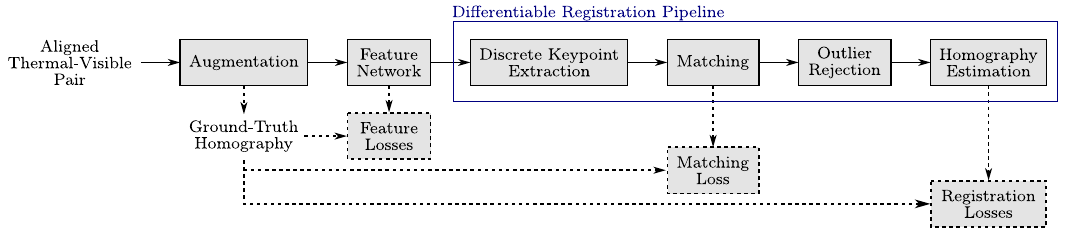}
	\caption{Training overview flowchart.
	One image of a thermal-visible image pair is transformed (i.e., augmented) by a random but known homography. 
	The feature network response to the augmented image pair is fed to a registration (i.e., homography estimation) pipeline. 
	Losses are applied to the network output, matches, and homography estimate.
	Dashed lines denote loss computation steps.
	}
	\label{fig:method-overview}
    \vspace{-6pt}
\end{figure*}

This section presents our approach to learning cross-spectral point features.
At a high-level, we feed our feature network's outputs into a differentiable registration (i.e., homography estimation) pipeline, and apply losses to the resulting matching and registration estimates.
An overview of our training framework is shown in Figure~\ref{fig:method-overview}.

Starting with an aligned thermal and visible-spectrum image pair,
we randomly sample a homography, $\HomGt$, and transform one of the images, simulating a viewpoint change from the \textit{source} (i.e., unchanged) image to the \textit{target} (i.e., transformed) image.
The transformed image pair is fed to the feature network, and from the network response, we extract and match point features, perform outlier rejection, and compute a homography estimate, $\HomEst$.
We apply losses to the matches and homography estimate, which are backpropagated through the differentiable pipeline to the model. 
We also apply losses to the network's detection and description outputs, similar to those in \cite{achermann_multipoint_2020}.
In addition, the differentiable registration pipeline, adapted from \cite{gridseth_keeping_2022}, contains no learned parameters, which constrains parameter updates to the feature network.

\subsection{Feature Network Architecture} 
\label{sec:network-architecture}

Our feature network architecture follows that of \cite{achermann_multipoint_2020}.
A VGG-style convolutional encoder \cite{simonyan_very_2015} compresses the input image to a latent representation of shape $H/8 \,\times\,W/8 \,\times\, 128$, where $H$ and $W$ are the height and width of the input images, respectively.
This branches into separate decoders for detection and description.
The descriptor decoder applies additional convolutions, followed by L2-normalization along the channel dimension to obtain a semi-dense descriptor map with $C$ channels, $\DescMap \in \Real^{H/8 \times W/8 \times C}$. 
The detector decoder applies additional convolutions to obtain an unnormalized detection response, $\LogitMap \in \Real^{H/8 \times W/8 \times 65}$, which is a collection of \textit{cells}, $\LogitCell \in \Real^{65}$. 
Each cell contains the keypoint response for an 8 $\times$ 8 pixel patch of the input image plus a \textit{dustbin} response for the absence of a keypoint.
The unnormalized detection response can be normalized into a keypoint probability heatmap by applying a cell-wise softmax, removing the now-redundant dustbins, and unpacking the remainder of the detection values into a probability heatmap of the original image size.
The same network is used for both spectra.

\subsection{Differentiable Registration Pipeline}
\label{sec:method-diff-reg-pipeline}

Our homography estimation pipeline builds on the differentiable pose estimator from \textcite{gridseth_keeping_2022}.
As mentioned earlier, the pipeline contains no learned parameters, which is intended to promote the generalization of the feature network \cite{gridseth_keeping_2022,sattler_understanding_2019}.

\subsubsection{Discrete Keypoint Extraction}
To extract discrete keypoints, 
we first unpack the unnormalized detection response into an array of size $H\times W$.
This array is divided into non-overlapping windows, 
and a subpixel keypoint coordinate, $\Point= \bbm u & v\ebm^T$, is estimated in each window using a spatial soft $\mathrm{argmax}(\cdot)$.
Low-quality detections are mitigated by assigning each keypoint a score, $\ScoreKP$, that is sampled from the (normalized) detection probability heatmap at the subpixel coordinate.
We also interpolate the keypoint's associated descriptor, $\Desc \in \Real^{C}$, from the descriptor map.

\subsubsection{Matching} 
For each source keypoint, $\PointSrc$, the differentiable matcher computes a \textit{pseudo-}target keypoint, $\PointTrgPseudo$, as a weighted sum of all extracted target keypoints.
This implicitly defines a match.
For source point $i$, $\PointSrc^{i}$, the pseudo-target keypoint location is computed as
\begin{equation}
   \PointTrgPseudo^{i} = \sum_{j=1}^{\NumTrgKp} {\SoftmaxElement\left(\frac{\fzncc(\DescSrc^i,\DescTrg^j)+1,}{\Temperature}\right)} \PointTrg^j ,
\end{equation}
where $\NumTrgKp$ is the number of target keypoints, 
$\fzncc(\cdot)$ denotes the \gls*{zncc} of two vectors, 
$\DescSrc^i$ and $\DescTrg^j$ are source descriptor $i$ and target descriptor $j$, respectively, 
and $\Temperature$ is the softmax temperature.
As before, the associated keypoint score, $\ScoreKPTargetPseudo$, and descriptor, $\DescTrgPseudo$, of the pseudo-target keypoint are interpolated from the detection probability heatmap and descriptor map, respectively.
Also, a matching score, $\ScoreMatch$, is given by
\begin{equation}
	\ScoreMatch=\frac{1}{2} ( \fzncc(\DescSrc,\DescTrgPseudo) + 1 ) , 
\end{equation}
which uses \gls*{zncc} to quantify the agreement of the matched descriptors and shifts the score range to $[0,1]$.

\subsubsection{Outlier Rejection} 
We reduce the impact of highly erroneous matches by assigning an inlier score, $\ScoreInlier$, to each match based on its reprojection error, $\MatchError$.
As in \cite{detone_superpoint_2018,achermann_multipoint_2020}, $\Hom\Point$ denotes the transformation of point $\Point$ by homography $\Hom$.
This score is calculated with an alternative sigmoid parametrization that starts close to unity for small reprojection errors and declines rapidly near an outlier threshold, $\OutlierThreshold$. The inlier score is computed as
\begin{equation}
	\ScoreInlier=\frac{1}{1+\exp\left(\OutlierSharpness \left(\MatchError/\OutlierThreshold - 1\right)\right)}, \quad \MatchError = \norm{{\HomGt}{\PointSrc} - {\PointTrgPseudo}}_2,
\end{equation}
where $\OutlierSharpness$ controls the score's rate of decline or \textit{sharpness}.
%

\subsubsection{Model Estimation}
\label{sec:method-model-estimation}
The estimated homography, $\HomEst$, is computed with the \acrlong*{dlt} algorithm, modified to weight each correspondence with the product of the keypoint, matching, and inlier scores, expressed as
\begin{equation} 
	\Score = {\ScoreKPSource}\:{\ScoreKPTargetPseudo}\:{\ScoreMatch}\:{\ScoreInlier}.
\end{equation}

\subsection{Losses}
\label{sec:method-losses}
For the task-based loss formulations, we consider two homography losses (corner and Frobenius norm) and a match loss (transfer).
We also compute detector and descriptor losses directly on the network outputs.
The total loss is a weighted sum of the corner ($\LossCorner$), Frobenius norm ($\LossFrob$), transfer ($\LossTrans$), descriptor ($\LossDesc$), and detector ($\LossDetector$) losses 
with respective weights $\LambdaCorner$, $\LambdaFrob$, $\LambdaTrans$, $\LambdaDesc$, and $\LambdaDetector$: 
\begin{equation}
	\Loss = \LambdaCorner\LossCorner + \LambdaFrob\LossFrob + \LambdaTrans\LossTrans + \LambdaDesc\LossDesc + \LambdaDetector\LossDetector.
\end{equation}
The relative weights are set empirically from a tuning stage, the results of which are presented in Sections~\ref{sec:results-reg-main} and \ref{sec:ablations}.

\subsubsection{Task-Based Loss}
\label{sec:reg-loss}

For each task-based loss, we first normalize all image coordinates to a range of $[-1,1]$
and then compute an error matrix in the forward (source-to-target) and inverse (target-to-source) directions, which creates a balanced loss.
These matrices are passed element-wise through a robust loss function, and then all elements are averaged.
For brevity, we will denote the normalized ground-truth and estimated homographies as $\HomGtNormSimp$ and $\HomEstNormSimp$, respectively.

\paragraph{Corner}
\label{sec:corner-loss}
The corner error matrix expresses the reprojection error of the (normalized) image corner coordinates, $\PointsCornerNorm$. 
We extend the notation of \cite{detone_superpoint_2018,achermann_multipoint_2020}
such that $\Hom\PointSet$ denotes the transformation of point set $\PointSet$ by homography $\Hom$.
The corner error matrices in the forward and inverse directions, $\ErrorCorner, \: \ErrorCorner\ErrorInv \in \Real^{2\times4}$, are expressed as
\begin{align}
	\begin{split}
	\label{eq:corner-error}
		\ErrorCorner          &= \PointsCornerNorm - \HomRes    \PointsCornerNorm , \\
		\ErrorCorner\ErrorInv &= \PointsCornerNorm - \HomResInv \PointsCornerNorm .
	\end{split}
\end{align}

\paragraph{Frobenius Norm}
\label{sec:frobenius-loss}
The Frobenius norm error, or Frobenius error for short, 
instead compares the parameters of the ground-truth and estimated homographies.
The forward and inverse Frobenius error matrices, $\ErrorFrob, \: \ErrorFrob\ErrorInv \in \Real^{3\times3}$, are defined as
\begin{equation}
	\ErrorFrob    =  \HomRes - \Identity_3, \quad 
	\ErrorFrob\ErrorInv = \HomResInv - \Identity_3, 
\end{equation}
where $\Identity_3$ is a 3$\times$3 identity matrix.

\paragraph{Transfer}
\label{sec:transfer-loss}
The transfer error matrices express the reprojection error of all $\NumMatch$ estimated correspondences when transformed into the same image with the ground truth homography.
The forward and inverse transfer error matrices, $\ErrorTrans, \: \ErrorTrans\ErrorInv \in \Real^{2 \times \NumMatch}$, are defined as
\begin{equation}
	\ErrorTrans = \HomGtNormSimp \PointSetSrcNorm - \PointSetTrgPseudoNorm, \quad
	\ErrorTrans\ErrorInv = \HomGtNormSimp^{-1} \PointSetTrgPseudoNorm - \PointSetSrcNorm,
\end{equation}
where $\PointSetSrcNorm$ and $\PointSetTrgPseudoNorm$ denote the sets of normalized source and pseudo-target keypoints, respectively.

\paragraph{Robust Loss Function}
We use Welsch loss \cite{dennis_jr_techniques_1978,barron_general_2019} for our robust loss function, which can be expressed as
\begin{equation}
	\frobust(\RobustError) = 1 - \exp\left(-\frac{1}{2}\left(\frac{\RobustError}{\RobustThreshold}\right)^2 \right) 
\end{equation}
for scalar error $\RobustError$, where $\RobustThreshold$ is a scaling hyperparameter.
For errors $\abs{\RobustError}<\RobustThreshold$, Welsch loss behaves similarly to squared-error loss, but the gradient diminishes past this point.

\subsubsection{Direct Feature Losses}
We draw on MultiPoint \cite{achermann_multipoint_2020} for our formulation of the detector and descriptor losses.

\paragraph{Descriptor}
The descriptor loss, $\LossDesc$, is a contrastive hinge loss between every source-target descriptor pair in the semi-dense descriptor maps.
For source descriptor $i$, $\DescSrc^i$, and target descriptor $j$, $\DescTrg^j$, the loss is expressed as
\begin{align}
	\begin{split}
	\LossDesc^{ij} = &\: (1-\Correspondence^{ij}) \max(0, {\DescCellSrc^i}^T\DescCellTrg^j - \MarginNeg) \\
					 &\; + \LambdaPos  \Correspondence^{ij} \max(0, \MarginPos - {\DescCellSrc^i}^T\DescCellTrg^j),
	\end{split}
\end{align}
where $\Correspondence^{ij}$ is a binary correspondence value, 
$\MarginPos$ and $\MarginNeg$ are the positive and negative hinge loss margins, respectively,
and $\LambdaPos$ is a weighting term to offset the low ratio of positive to negative correspondences.
The overall descriptor loss is averaged over all source-target descriptor pairs.

\paragraph{Detector}
\label{sec:detector-loss}
In \cite{achermann_multipoint_2020}, the detector loss is formulated as a per-cell classification problem and uses the keypoint pseudo-ground truth provided with their dataset. 
We add a weighting vector, $\DetLossWeight$, to the categorical cross-entropy loss in order to downweight the overrepresented ``no keypoint'' case.
For cell $i$, $\LogitCell^{i}$, the loss is expressed as
\begin{equation}
	\LossDetector^{i} =  - \sum_{j=1}^{65} \DetLossWeightElement{j} \OneHotElement{j}^{i} \log\left(\frac{\exp(\LogitCellElement{j}^{i})}{\sum_{l=1}^{65} \exp(\LogitCellElement{l}^{i})} \right) , 
\end{equation}
where $\OneHotVector^{i}$ is the one-hot encoded label.
The final detector loss, $\LossDetector$, is averaged over all cells in the source and target images.


\subsection{Implementation Details}

Our feature network uses descriptors of length $C=64$, matching that of \cite{achermann_multipoint_2020}.
In the differentiable registration pipeline, 
we use keypoint extraction windows of size 8 $\times$ 8 pixels;
a temperature value of $\tau=0.01$ for pseudo-target keypoint computation;  
and for a training image size of 240 $\times$ 320 pixels, we use a rejection threshold of $\OutlierThreshold=50$ pixels and a sharpness parameter of $\OutlierSharpness=5$. 
We use a robust threshold of $\RobustThreshold=0.1$ on all task-based errors.
For the descriptor loss, we use a positive margin of $\MarginPos=1.0$, a negative margin of $\MarginNeg=0.2$, and a positive correspondence weighting of $\LambdaPos = 250$ as in \cite{achermann_multipoint_2020}.
We choose detection loss weightings of $\DetLossWeightElement{j}=64/65$ for $j=1,\hdots,64$ and $\DetLossWeightElement{65}=1/65$ for the ``no keypoint'' case.

\section{Experiments}

In this section, we describe our experimental setup to evaluate the performance of our model on cross-spectral feature detection, matching, and homography estimation.

\subsection{Estimation Framework}
Much like our training framework, the testing framework begins with an aligned thermal-visible image pair. We sample and apply the ground-truth homography, $\HomGt$, pass the transformed images to the feature algorithm, and feed the feature response into a registration pipeline to obtain $\HomEst$. 

We use two different pipelines for evaluation. 
For all methods, we use a `classical' pipeline that executes a standard feature-based registration procedure, including mutual nearest neighbour matching, \gls*{ransac} outlier rejection, and damped least-squares for model estimation.
When using learned methods on the classical pipeline, we apply a detection threshold followed by \acrlong*{nms} to extract discrete keypoints as in \cite{achermann_multipoint_2020}.

We also use a `weighted' pipeline, tailored to learned features, that contains the same keypoint extraction, matching, and model estimation blocks as the training pipeline.
For outlier rejection, the weighted pipeline instead uses \gls*{ransac}, modified to weight the random sampling with the keypoint and match scores.
Inlier matches receive an inlier score of one ($\ScoreInlier^{i}=1$) and zero otherwise.

\subsection{Performance Metrics}
Our registration and standalone feature performance metrics are computed as follows. 

\begin{figure}[]
	\vspace{6pt}
    \centering
    \subfloat[ACE=5  ]{\includegraphics[width=1.1in]{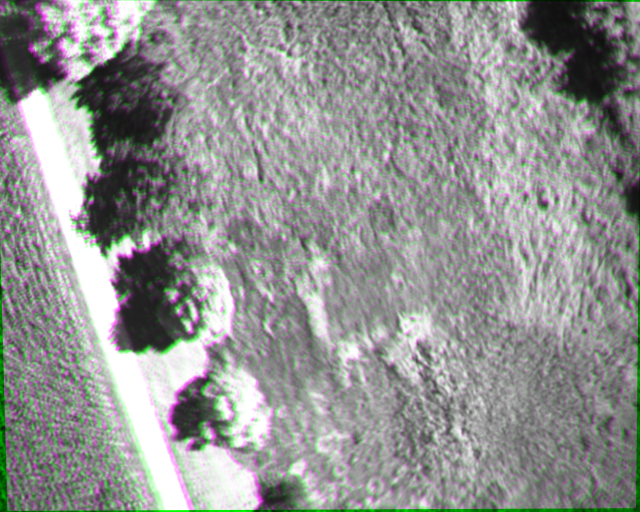}} \hfill
    \subfloat[ACE=20 ]{\includegraphics[width=1.1in]{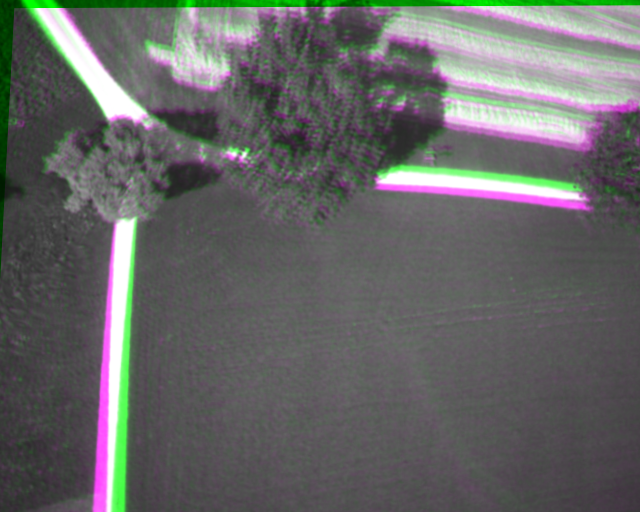}} \hfill
    \subfloat[ACE=100]{\includegraphics[width=1.1in]{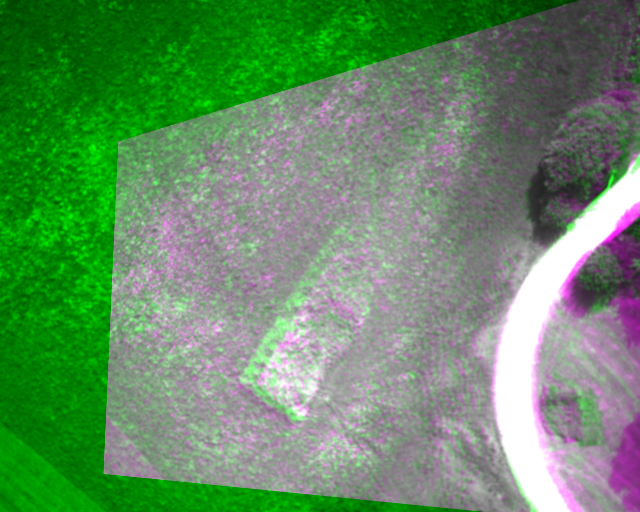}} \\
    \caption{Alignment visualization for different \acrfull*{ace} values (expressed in units of pixels) on images of shape 512 $\times$ 640. Each panel shows a rectified image in green overlaid with its ``recovered’’ version in purple.}
    \label{fig:ace-examples}
    \vspace{-6pt}
\end{figure}

\paragraph{Registration}
For each image pair, we compute the \gls*{ace}, 
defined as the average reprojection error of the image corners 
when transformed first by the ground-truth homography followed by the inverse estimated homography, expressed as
\begin{align}
	\begin{split}
    \ACE = \frac{1}{4} \norm{ \PointsCornerFull - ({\HomEst}^{-1} \HomGt)\:\PointsCornerFull}_2, 
   \label{eq:average-corner-error}
\end{split}
\end{align}
where $\PointsCornerFull$ is the set of (unnormalized) image corner coordinates. Recall that $\Hom\PointSet \in \Real^{2\times N}$ denotes the transformation of a point set.
Figure \ref{fig:ace-examples} shows alignment attempts at varying \gls*{ace} levels.
Over the test set, we report the proportion of estimates with \gls*{ace} below different pixel thresholds, $\ACEThresh=\set{2,5,10,25}$, for images of shape 512 $\times$ 640. 
%

\paragraph{Features}
Following the evaluation procedure of \cite{detone_superpoint_2018,achermann_multipoint_2020}, we compute the repeatability and \gls*{mscore} for each image pair and report the per-image average over the test set. 
We also compute \gls*{mma} with the same procedure.
In addition, we calculate the \gls*{map} over the dataset.
Repeatability is the ratio of keypoints detected at the same physical locations in both images to the overall number of detections.
\Gls*{mscore} reports the ratio of correct matches to total detections. 
\Gls*{mma} and \gls*{map} measure the descriptor's discriminating power: \gls*{mma} is the ratio of correct to reported matches while \gls*{map} reports the area under the precision-recall curve.
We additionally report the average number of detections per image, $\NumKp$. 

\subsection{Dataset}
\label{sec:dataset}

For training and testing, we use the MultiPoint dataset \cite{achermann_multipoint_2020}, which consists of aligned pairs of thermal and visible-spectrum images captured by a \gls*{uav} over an agricultural area.
Additionally, this dataset provides keypoint pseudo-ground truth generated from multi-spectral homographic adaptation.
Data was collected over ten flights under varying lighting conditions and large temperature changes. 
From these potentially overlapping flight paths, seven flights were used for training data and three for testing, yielding 9340 and 4391 image pairs, respectively. 
During training, images are randomly cropped to a size of 240 $\times$ 320 pixels and undergo photometric augmentation in addition to geometric transformation.
%
\subsection{Baselines}
\label{sec:baselines}
We consider many categories of feature algorithms and compare against the standard methods for each,
including two handcrafted visible-spectrum methods, \gls*{sift} \cite{lowe_distinctive_2004} and \gls*{orb} \cite{rublee_orb_2011}; a learned visible-spectrum method, SuperPoint \cite{detone_superpoint_2018}; a handcrafted cross-spectral method, the \gls*{lghd} \cite{aguilera_lghd_2015}; and a learned cross-spectral method, MultiPoint \cite{achermann_multipoint_2020}.
We use the Python implementation of \gls*{lghd} from \cite{achermann_multipoint_2020}, which pairs \gls*{lghd} with the \gls*{fast} detector.
For SuperPoint, we use the publicly available weights. We retrained MultiPoint after splitting the training set into training (80\%) and validation (20\%) subsets.
We also trained MultiPoint-W, a version of MultiPoint that uses our weighted detector loss in place of the original, unweighted loss.

\subsection{Training Details}
\label{sec:results-training}

Of the task-based losses, we opt to use \textit{only} the transfer loss with a weighting of $\LambdaTrans=1$ in our final model.
Our task-based loss comparison is discussed in Section~\ref{sec:ablations}.
We also use the descriptor loss and weighted detector loss ($\LambdaDesc = \LambdaDetector = 1$), subsequently referred to as the `base' losses.
The base losses are used to train MultiPoint-W, which we use to initialize our weights.
We train with the PyTorch package \cite{paszke_pytorch_2019} and the Adam optimizer \cite{kingma_adam_2015} with learning rate $10^{-5}$ and batch size $32$. 
Although we implement an early stopping mechanism to halt training when our model's registration performance on the validation set plateaus, this does not trigger, and our model trains for the full 1000 epochs.
We also do not observe an increase in validation loss, indicating no signs of overfitting to the training data.
Finally, we extend the training of MultiPoint and MultiPoint-W for the same number of epochs at the same learning rate as our model.

\section{Results and Discussion}
In this section, we present the evaluation results of our feature model against a set of baseline methods and compare the effectiveness of the task-based losses.

\subsection{Registration Performance}
\label{sec:results-reg-main}

We perform homography estimation with all methods on the classical pipeline and learned methods on the weighted pipeline.
Table~\ref{tab:reg-all} summarizes the registration performance results.
As expected, visible-spectrum methods, whether handcrafted (\gls*{sift} and \gls*{orb}) or learned (SuperPoint), obtain low registration success rates. Of these methods, \gls*{sift} performs the best, with 21\% of estimates under 5 pixels of error.
The handcrafted cross-spectral method, \gls*{lghd}, 
also demonstrates a low registration success rate on this dataset, with only 6\% of estimates below 25 pixels of error.
As noted by \cite{achermann_multipoint_2020}, this is likely due to its lack of viewpoint invariance.
Learned methods may be better equipped than handcrafted algorithms to achieve simultaneous invariance to dramatic appearance and geometric changes, provided that representative data is available.
Indeed, all cross-spectral learned methods (MultiPoint, MultiPoint-W, and our model) achieve an \gls*{ace} of less than 5 pixels for almost half of the estimates.
Our model on the weighted pipeline achieves the highest success rates, obtaining an \gls*{ace} of less than 10 pixels for over 75\% of all estimates.
\begin{table}[h!]
    \centering
        \caption{
        Registration performance for all pipeline-method combinations, 
        measured by proportion of estimates for which $\ACE < \ACEThresh$ pixels.
        The best result for each metric is bolded.
        }
        \label{tab:reg-all}
    \begin{tabular}{@{}llllll@{}}
    \toprule
    Pipeline                   & Method                                      & $\ACEThresh=2$   & $\ACEThresh=5$   & $\ACEThresh=10$  & $\ACEThresh=25$  \\ \midrule
    \multirow{7}{*}{Classical} & SIFT \cite{lowe_distinctive_2004}           & 0.0608         & 0.210          & 0.276          & 0.322          \\
                               & ORB \cite{rublee_orb_2011}                  & 0.000911       & 0.0100         & 0.0339         & 0.0854         \\
                               & SuperPoint \cite{detone_superpoint_2018}    & 0.0137         & 0.0847         & 0.135          & 0.175          \\
                               & LGHD \cite{aguilera_lghd_2015}              & 0.00547        & 0.0269         & 0.0455         & 0.0599         \\
                               & MultiPoint \cite{achermann_multipoint_2020} & 0.122          & 0.494          & 0.647          & 0.716          \\
                               & MultiPoint-W                                & 0.134          & 0.536          & 0.675          & 0.750          \\
                               & Ours                                        & 0.120          & 0.569          & 0.713          & 0.781          \\ \midrule
    \multirow{4}{*}{Weighted}  & SuperPoint                                  & 0.0146         & 0.0843         & 0.143          & 0.195          \\
                               & MultiPoint                                  & 0.107          & 0.516          & 0.671          & 0.728          \\
                               & MultiPoint-W                                & 0.158          & 0.573          & 0.718          & 0.786          \\
                               & Ours                                        & \textbf{0.197} & \textbf{0.614} & \textbf{0.760} & \textbf{0.824} \\ \bottomrule
    \end{tabular}
    \vspace{-6pt}
    \end{table}

\begin{figure}[b!]
    \centering
    \includegraphics[width=3.35in]{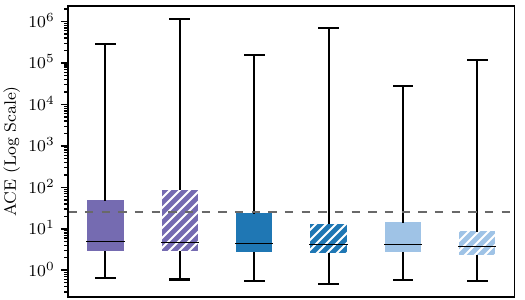}
    \vspace{-1mm}
    \includegraphics[width=3.35in]{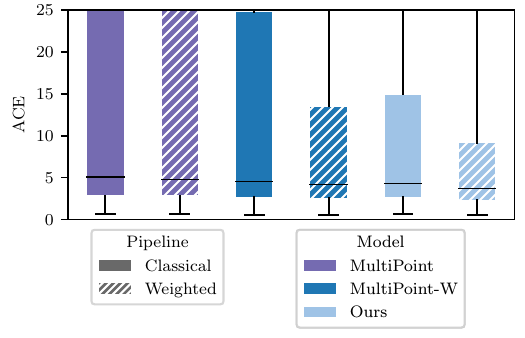}
    \caption{Box plots of \acrfull*{ace} expressed in pixels in logarithmic (top) and linear (bottom) scale for the cross-spectral learned feature methods on both pipelines. The whiskers extend to all data points. 
    }
    \label{fig:box-mp-variants}
\end{figure}

Focusing on the cross-spectral learned methods,
Figure~\ref{fig:box-mp-variants} shows \gls*{ace} box plots for the cross-spectral learned features on both pipelines: 
the top plot shows the entire error distributions on a logarithmic scale (as homography errors can span many orders of magnitude) and the bottom plot shows errors below 25 pixels on a linear scale.
Notably, training our model on the differentiable registration pipeline improves the performance of our model on the nondifferentiable, \textit{classical} pipeline despite the differences between the two pipelines.
In addition, the weighted detector loss appears to improve the registration performance of the MultiPoint network with both pipelines.
These observations are further discussed in Section~\ref{sec:results-features}.

\subsection{Standalone Feature Performance}
\label{sec:results-features}

We compute the feature detection and matching metrics for all methods on the classical pipeline.
These metrics are not computed on the weighted pipeline as it does not count all detections and matches equally.
Table~\ref{tab:feat-classical} summarizes the results.
Our method achieves the highest repeatability, \gls*{mscore}, and \gls*{mma} on the classical pipeline despite its dissimilarity from the training pipeline.
This results validates our decision to use a differentiable registration pipeline without learned parameters.

\begin{table}[h!]
\centering
    \caption{Feature performance metrics for all methods on the classical pipeline. 
    The best result for each applicable metric is bolded.}
    \label{tab:feat-classical}
\begin{tabular}{@{}lrllll@{}}
    \toprule
        Method                                      & $\NumKp$ & Rep.  & \acrshort*{mscore}  & \acrshort*{mma}     & \acrshort*{map}     \\ \midrule
        SIFT \cite{lowe_distinctive_2004}           & 643      & 0.269          & 0.0294         & 0.103          & 0.0223         \\
        ORB \cite{rublee_orb_2011}                  & 359      & 0.273          & 0.0188         & 0.0572         & 0.0120         \\
        SuperPoint \cite{detone_superpoint_2018}    & 248      & 0.183          & 0.0274         & 0.0876         & 0.0664         \\
        LGHD \cite{aguilera_lghd_2015}              & 1243     & 0.234          & 0.00448        & 0.0371         & 0.00564        \\
        MultiPoint \cite{achermann_multipoint_2020} & 438      & 0.291          & 0.111          & 0.295          & \textbf{0.206} \\
        MultiPoint-W                                & 1609     & 0.446          & 0.107          & 0.278          & 0.150          \\
        Ours                                        & 1708     & \textbf{0.454} & \textbf{0.124} & \textbf{0.317} & 0.189          \\ \bottomrule
    \end{tabular}
\vspace{-6pt}
\end{table}

We further consider how the number of keypoints and weighted detector loss influence these results. 
For a given feature method, an increased number of keypoints is positively correlated with repeatability and often negatively correlated with descriptor-based metrics due to the higher risk of false positives.
Our model and MultiPoint-W exhibit higher repeatability scores than the original MultiPoint by a wide margin, improving from 29\% to 45\%. 
This appears to be due, in part, to the promotion of keypoints from the weighted detector loss.
If we artificially increase the number of MultiPoint’s detections, as shown in Table~\ref{tab:feat-mp-threshold}, we observe that a similar number of keypoints yields comparable repeatability to our method and MultiPoint-W, but with lower descriptor metrics.
Therefore, it appears that the weighted detector loss helps our method and MultiPoint-W achieve high levels of repeatability without sacrificing matching accuracy, and our method receives an additional performance boost from our task-oriented training.

\begin{table}[h]
\centering
    \caption{MultiPoint network’s feature performance metrics with the classical pipeline at varying detection thresholds.
    The best result for each applicable metric is bolded. 
    A lower detection threshold increases the repeatability but diminishes the matching performance.}
    \label{tab:feat-mp-threshold}
\begin{tabular}{@{}lrllll@{}}
    \toprule
    Detection Threshold                      & $\NumKp$ & Rep.  & \acrshort*{mscore} & \acrshort*{mma} & \acrshort*{map} \\ \midrule
    1.5$\times 10^{-2}$ (Original) & 438      & 0.291          & \textbf{0.111}     & \textbf{0.295}  & \textbf{0.206}  \\
    5.0$\times 10^{-3}$              & 910      & 0.369          & 0.0995             & 0.271           & 0.150           \\
    1.5$\times 10^{-3}$            & 1731     & \textbf{0.460} & 0.0854             & 0.254           & 0.102           \\ \bottomrule
\end{tabular}
\vspace{-6pt}
\end{table}

\subsection{Task-Based Loss Comparison}
\label{sec:ablations}

As mentioned in Section~\ref{sec:results-training}, 
we use only transfer loss out of the task-based losses.
We also train model variants that either substitute for or add in a homography-based loss (corner or Frobenius). 
Table~\ref{tab:ablation} shows the registration results of all model variants with the weighted pipeline.
We observe that models trained with the homography losses have lower success rates than their counterparts without these losses.

Further investigation reveals that homography-based loss functions are incompatible with our specific training pipeline. 
In our framework, the homography errors are backpropagated through the outlier rejection block to the match locations. 
However, the ground-truth outlier rejection does not impose constraints on the matches required to obtain a unique solution, creating an ill-posed optimization problem.
As a result, in the training pipeline, we observe an ``averaging’’ effect in which the match locations are adjusted to offset each other’s errors.
Our findings agree with \textcite{gridseth_keeping_2022}, who use ground-truth outlier rejection in their training pipeline and found that a matching loss was needed in addition to pose error.
Therefore, task-based losses can be effective provided that they are integrated into a cohesive geometric framework.
\begin{table}[t]
\centering
    \caption{
        Registration performance for models trained with different loss combinations,
        measured by rate of estimates for which $\ACE < \ACEThresh$ pixels.
        The best result for each metric is bolded. 
        Models using homography-based losses (corner or Frobenius) 
        perform worse than those trained without.
    }
    \label{tab:ablation}
\begin{tabular}{@{}lrlll@{}}
\toprule
    Losses                  & $\ACEThresh=2$   & $\ACEThresh=5$   & $\ACEThresh=10$  & $\ACEThresh=25$  \\ \midrule
    Base (MultiPoint-W)     & 0.158          & 0.573          & 0.718          & 0.786          \\
    Base+Transfer (Ours)    & \textbf{0.197} & \textbf{0.614} & \textbf{0.760} & \textbf{0.824} \\
    Base+Corner             & 0.137          & 0.517          & 0.668          & 0.745          \\
    Base+Frobenius          & 0.144          & 0.539          & 0.686          & 0.763          \\
    Base+Transfer+Corner    & 0.171          & 0.590          & 0.735          & 0.803          \\
    Base+Transfer+Frobenius & 0.179          & 0.603          & 0.747          & 0.808          \\
    \bottomrule
    \vspace{-6pt}
\end{tabular}
\end{table}

\section{Conclusion}
In this paper, we presented our task-oriented approach to learning cross-spectral point features.
In addition to standard detection and description losses,
we trained our feature network on its matching performance 
by using a non-learned, differentiable registration pipeline. 
Our method achieved lower registration errors than a set of baselines on the MultiPoint dataset \cite{achermann_multipoint_2020},
obtaining a registration error of less than 10 pixels for more than 75\% of estimates.
We also demonstrated improved matching and registration performance compared to the baselines with a classical registration pipeline, despite its dissimilarity from the differentiable registration pipeline used during training.
We further examined the relative effectiveness of task-based losses and explored why homography-based losses were not compatible with our specific training framework.
A promising avenue for future work involves incorporating differentiable \gls*{ransac} into the training pipeline, 
which could introduce the geometric constraints needed to train on the registration error.

\balance
\printbibliography

\end{document}